\documentclass[11pt,a4paper]{article}
\usepackage[hyperref]{emnlp2020}
\usepackage{times}
\usepackage{latexsym}

\usepackage{tabularx}
\usepackage{amsmath}
\usepackage{amsfonts}
\usepackage{enumitem}
\usepackage{url}
\usepackage{graphicx, wrapfig, subcaption, setspace, booktabs}

\usepackage{microtype}

\aclfinalcopy 

\title{Human brain activity for machine attention}

\author{Lukas Muttenthaler$^{\dagger}$~Nora Hollenstein$^{\ddagger}$~Maria Barrett$^{\dagger \odot}$ \\
  $^{\dagger}$ Dept. of Computer Science, University of Copenhagen \\
  $^{\ddagger}$ Dept. of Computer Science, ETH Zurich \\
  $^{\odot}$ Dept. of Computer Science, IT University of Copenhagen \\
  \texttt{mnd926@alumni.ku.dk, noraho@inf.ethz.ch, mjb@di.ku.dk}}

\date{June 2020}

\begin{document}
\maketitle
\begin{abstract}
Cognitively inspired NLP leverages human-derived data to teach machines about language processing mechanisms. Recently,  neural networks have been augmented with behavioral data to solve a range of NLP tasks spanning syntax and semantics. 
We are the first to exploit neuroscientific data, namely electroencephalography (EEG), to inform a neural attention model about language processing of the human brain. 
The challenge in working with EEG data is that features are exceptionally rich and need extensive pre-processing to isolate signals specific to text processing. 
We devise a method for finding such EEG features to supervise machine attention through combining theoretically motivated cropping with random forest tree splits. 
After this dimensionality reduction, the pre-processed EEG features are capable of distinguishing two reading tasks retrieved from a publicly available EEG corpus.
We apply these features to regularise attention on relation classification, and show that EEG is more informative than strong baselines. This improvement depends on both the cognitive load of the task and the EEG frequency domain.
Hence, informing neural attention models with EEG signals is beneficial but requires further investigation to understand which dimensions are the most useful across NLP tasks.
\end{abstract}


\graphicspath{ {./images/} }
\newcommand{\HRule}[1]{\rule{\linewidth}{#1}}

\section{Introduction}
Cognitively inspired NLP is a research field at the intersection of Cognitive Neuroscience and Natural Language Processing (NLP), which lately has received growing attention. Cognitive Neuroscience aims to investigate cognitive processes that occur in the human brain through high-level explanations, whereas NLP has the overall objective to teach machines to read and understand human language. The recent merge of these fields has led to the overarching goal of introducing human bias \cite{wilson2015human,DBLP:conf/nips/SchwartzTW19,DBLP:conf/nips/TonevaW19} into machines, and hence augment neural networks with cognitive data in solving NLP tasks. 

Human readers process words without notable effort.
Human text processing can be studied e.g. through eye tracking (ET) records from reading where fixation durations on word level are robustly correlated with cognitive load \cite{drieghe2005eye,fitzsimmons2011influence,rayner2011eye}. Recently, numerous studies have proven that words represented as ET features can help a wide range of NLP tasks including part-of-speech tagging \cite{barrett2016weakly}, relation detection \cite{hollenstein2019advancing}, and sentiment analysis \cite{mishra2017learning,mishra2017leveraging,barrett2018sequence}. 

Fixation durations reliably correlate with cognitive load \cite{rayner2011eye} but prolonged fixation duration will not help differentiate which cognitive process occurs. Therefore, eye movements are considered indirect measures of human text processing whereas electroencephalography (EEG) and fMRI technologies are direct measures of human brain activity. EEG measures electric activity in the brain. When used non-intrusively, a number of electrodes are placed on the scalp to measure brain surface activity. In the field of Cognitive Science, EEG data plays a vital role to explain various phenomena such as cognitive load \cite{antonenko2010using,zarjam2011spectral,kumar2016measurement}. It is even possible to decode human cortical activity to synthesize audible speech \cite{anumanchipalli2019speech}. 

In NLP, however, there has just been a single study that investigated how EEG data can enhance machines' ability to perform named entity recognition, sentiment, and relation classification \cite{hollenstein2019advancing}. 
In contrast to eye movements, EEG contains not only signals about cognitive load and mental state, but \textit{all} brain activity within the temporal domain (including facial muscle activity which must be removed from the data after recording). Therefore, EEG requires considerably more de-noising and pre-processing steps before any text processing signals are exploitable for NLP. 

Providing a rigorous and clear approach for the latter is the main aim of this study. On top of that, we go one step further and show how our approach improves performance on three sequence classification tasks - even with access to limited amounts of brain data. 

\subsection{Contributions}

\begin{itemize}
    \item We devise a method for extracting human language processing signals from EEG recordings of a publicly available corpus \cite{hollenstein2018zuco}. As a sanity check, we demonstrate that these multi-dimensional feature vectors let us distinguish between two different reading tasks, namely Normal Reading (NR) and Annotation Reading (AR).
    \item  We inject this human bias into a multi-task neural model for sequence classification. In doing so, we use the most informative EEG signals on the word-level to regularize the attention mechanism of an RNN.
    \item We observe that the differences in EEG signals from NR and AR affect NLP downstream performances differently. Moreover, we show that downstream performance further varies as a function of EEG frequency domain.
\end{itemize}

Together, these insights have decisive implications about which human EEG signals are most suitable to inject into Machine Learning (ML) models for enhancing their language processing performance. We make our code publicly available. \footnote{\url{https://github.com/LukasMut/NER-with-EEG-and-ET}}


\section{Related work}
Recently, a number of studies has investigated how external cognitive signals, and thus the injection of human bias, can enhance the capacity of artificial neural networks to understand natural language \cite{hollenstein2019advancing,strzyz2019towards,DBLP:conf/nips/SchwartzTW19,DBLP:conf/nips/TonevaW19,DBLP:conf/emnlp/GauthierL19}, and vice versa, how language processing in neural networks might enhance our understanding of human language processing \cite{hollenstein-etal-2019-cognival}. Others scrutinized whether machine attention deviates from human attention when disentangling visual scenes \cite{das2017human}.


Most studies, however, focused on the use of ET data, and hence exploited gaze features on the word level. Some utilized gaze features as word embeddings to inform neural networks about which syntactic linguistic features humans deem decisive in their language processing. In so doing, they have successfully enhanced state-of-the-art models for named entity recognition (NER; \newcite{hollenstein2019entity}), part-of-speech tagging \cite{barrett2016weakly} and dependency parsing \cite{strzyz2019towards}. Others have drawn attention to the enhancement of semantic disentanglement, and improved tasks such as sarcasm detection \cite{mishra2017learning}, or sentiment analysis \cite{mishra2017leveraging} through leveraging human gaze.


One recent study, from which we take inspiration, exploited gaze information to regularize attention in a multi-task-like setting for sequence classification \cite{barrett2018sequence}. Here, gaze information improved grammatical error detection, hate speech detection, and sentiment classification. The authors enabled neural networks to utilize human attention during training time without the need of accessing this information during test time. Although this study is similar to ours and serves as the foundation for our code, we go one step beyond, and regularize attention with human brain activity instead of indirect proxies such as gaze.

\section{EEG feature extraction}
In this section, we introduce the EEG corpus used for this study, then explain our feature extraction and dimensionality reduction approach in detail.

\subsection{EEG data}\label{sec:data}

In this study, we use the recently created Zurich Cognitive Language Processing Corpus (ZuCo)\footnote{\url{https://osf.io/q3zws/}} \cite{hollenstein2018zuco}. It is a corpus of simultaneous ET and EEG recordings of 12 English native speakers reading individual sentences. Thus, through the ET and EEG alignment, neural activities are available for single words in English. The corpus contains cognitive data from the following three reading tasks: 

\paragraph{Task 1} Normal Reading (NR) of 400 sentences that contain sentiment annotations from the Stanford Sentiment Treebank \cite{socher2013recursive}. In NR, reading comprehension was the sole task. 
\paragraph{Task 2} NR of 300 sentences that contain relations between named entities extracted from the Wikipedia relation extraction corpus \cite{culotta2006integrating}.
\paragraph{Task 3} Annotation Reading (AR) of 407 sentences that contain relations between named entities, also from \citet{culotta2006integrating}. Participants were required to annotate sentences while reading by pressing a key to answer whether one specific relation is present in the sentence. This results in a different cognitive load for the human reader.\\

Each of the 12 participants had to perform all three reading tasks. Participants completed the readings in two sessions on different days. For our experiments, we exclusively used data from NR (Task 2) and AR since these tasks contain text from the same domain. 


EEG features are recorded simultaneously with a corresponding ET feature (e.g., First Fixation Duration). As such, eight 105-dimensional EEG vectors were extracted for each ET feature corresponding to the EEG activity during the specific eye-movement event (i.e., ET feature) w.r.t. this word. Each of the eight 105-dimensional vectors corresponds to one of the following EEG frequency domains: $\theta_{1}$ (4-6 Hz), $\theta_{2}$ (6.5-8 Hz) $\alpha_{1}$ (8.5-10 Hz), $\alpha_{2}$ (10.5-13 Hz), $\beta_{1}$ (13.5-18 Hz) $\beta_{2}$ (18.5-30 Hz), $\gamma_{1}$ (30.5-40 Hz) and $\gamma_{2}$ (40-49.5 Hz) \cite{hollenstein2018zuco}. Each of the 105 dimensions corresponds to a specific electrode on a 128-channel EEG cap used for recording (see Figure~\ref{fig:Figure_1}). 23 electrodes were removed through Automagic \cite{pedroni2019automagic} prior to any data analysis, since they did not contain signals relevant for cognitive processing due to their position \cite{hollenstein2018zuco}.

Given that $n$ = number of recorded ET features, $m$ = number of EEG frequency domains, and, $k$ = number of electrodes, this results in $n \times m \times k = 4200$ EEG features per word per ET feature, where $n = 5$, $m = 8$, and, $k= 105$. Since most of these $4200$ features reflect signals that are not relevant to cognitive text processing, we apply various feature extraction techniques to reduce the dimensionality.

\subsection{EEG feature reduction}
We split the EEG data for NR and AR into a train, development and test split with 70/15/15\% of the data. Splits were created w.r.t. sentence-level samples. We calculated feature reduction on the train set, tuned it on development splits, and evaluated the reduced features on a held-out test set. 

We first investigated how to pre-process the EEG features to find differences between AR and NR.
This was done to scrutinize potential differences in EEG signals as a function of cognitive load, and evaluate whether the dimensionality reduction is able to capture those. As a validation step, we also tuned the method for classifying Task 1 into its first and second halves respectively. Participants completed Task 2 and the first half of Task 1 in the first session, and Task 3 and the second half of Task 2 in the second session. Such experimental designs may bias the data with session-specific effects. Hence, this validation step serves both as a data quality check of the dimensionality reduction as well as of the noise removal. Ideally, the text processing signal should remain coherent for both halves of Task 1. For both steps, we employed a Random Forest classifier to the train set to find the most informative EEG channels for this task.

\begin{figure*}[ht]
\centering
\begin{subfigure}{.33\textwidth}
    \centering
    \includegraphics[width=0.95\textwidth]{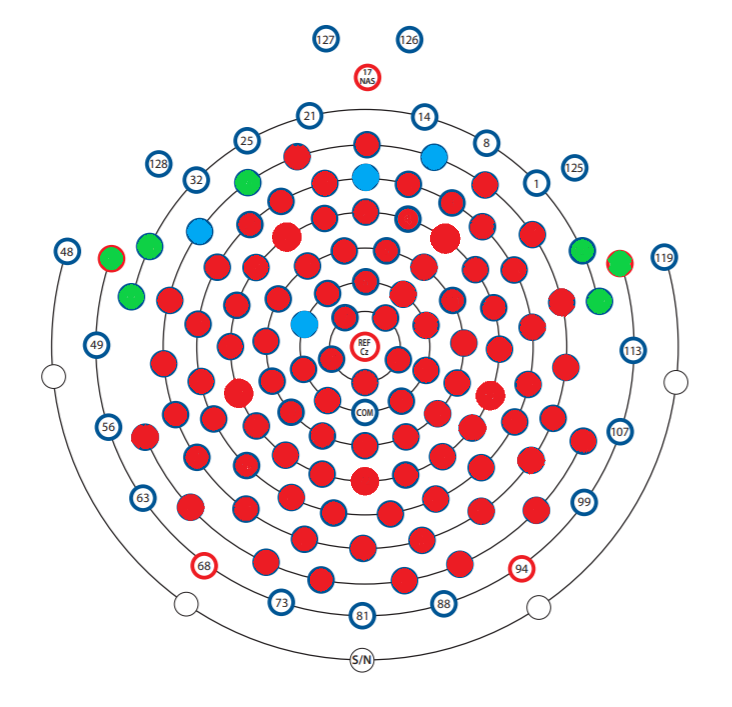}
    \caption{$\theta$-frequency band (4-8 Hz)}
\end{subfigure}%
\begin{subfigure}{.33\textwidth}
    \centering
    \includegraphics[width=0.95\textwidth]{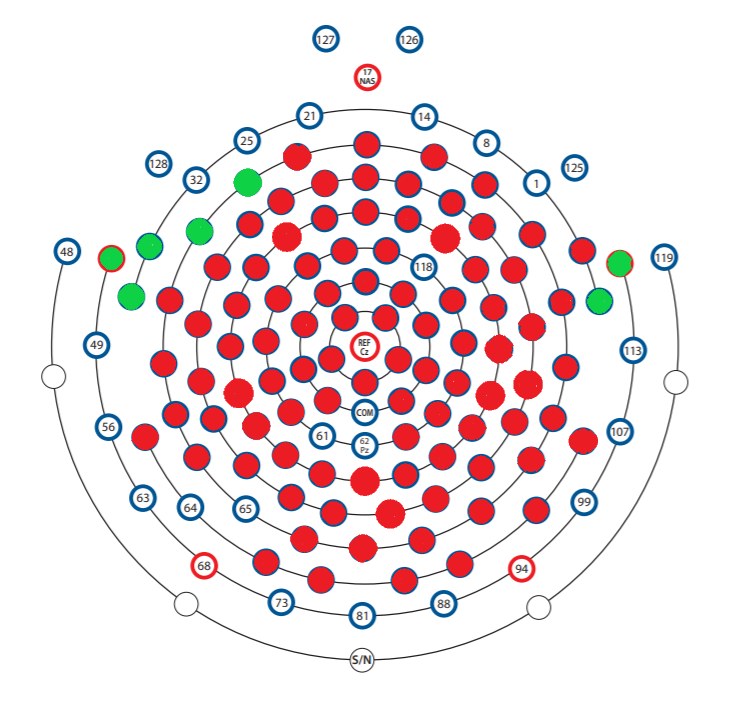}
    \caption{$\alpha$-frequency band (8.5-13 Hz)}
\end{subfigure}
\begin{subfigure}{.33\textwidth}
    \centering
    \includegraphics[width=0.95\textwidth]{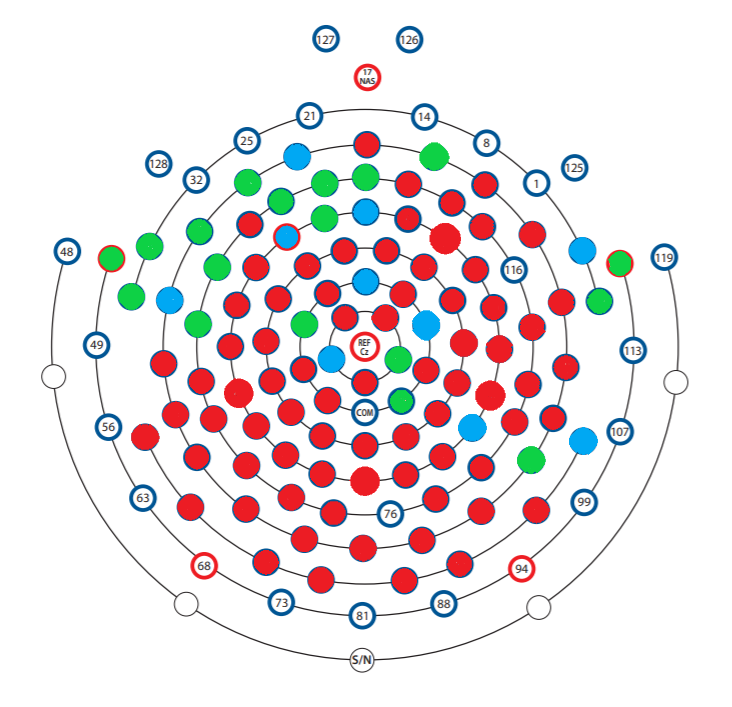}
    \caption{$\beta$-frequency band (13.5-30 Hz)}
\end{subfigure}%
\caption{Differences in EEG power spectra for TRT in specific electrodes between Normal Reading (NR) and Annotation Reading (AR) for the four frequency bands $\theta$, $\alpha$, $\beta$. Best viewed in color. Electrodes colored in green denote higher EEG power spectra for AR, whereas red electrodes indicate higher EEG signals for NR. All differences are statistically significant with $p < 0.01$. Blue colored electrodes refer to no or non-significant differences between the two reading tasks. All electrodes that are not colored are electrodes that have been excluded during pre-processing prior to data analysis. The reference electrode (Cz), however, was not excluded as a pre-processing step, but not considered during statistical analyses since it always has the minimum value of 0.}
\label{fig:Figure_1}
\end{figure*}

Firstly, to extract the $k$ most predictive features per word, per ET feature, and per EEG frequency domain we reduced $m$ and $n$ based on literature.

\paragraph{Reducing $m$}
 We binned the eight frequency domains (see above) into the four general frequency bands, $\theta$ (4-8 Hz), $\alpha$ (8.5-13 Hz), $\beta$ (13.5-30 Hz) and $\gamma$ (30.5-49.5 Hz). This strategy was applied to manually decrease the dimensionality prior to exploiting any machine learning techniques, and reduce computational cost at an early stage. To yield binned frequency domains, we calculated the average power spectrum per electrode for each of the four frequency pairs (e.g., $\bar x(\alpha_1, \alpha_2)$), thus $m=4$. Due to the fact that $\gamma$ frequencies mainly relate to emotionality \cite{zheng2015investigating,li2009emotion,oathes2008worry,luo2008visual}, and not to attentiveness, we further reduced $m = 4$ to $m = 3$, and computed embeddings in low-dimensional brain space for $\alpha$, $\beta$, and $\theta$ only.
 
 \paragraph{Reducing $n$}
 In initial data analyses, kernel density estimates (KDE) and $t$-test bootstrapping showed that power spectra vary across EEG frequency domains but are highly correlated among the different ET features. Therefore, we decided to extract EEG features that correspond to total reading time (TRT), as this ET feature covers all activity related to an individual word, and has proven to be the most informative gaze feature in previous studies \cite{hollenstein2019entity,hollenstein2019advancing}. Hence, $n=1$.

Furthermore, we exploited a Random Forest classifier \cite{breiman2001random} with 100 trees. 
A random forest \footnote{Scikit learn \cite{scikit-learn} with default parameters. However, we set the bootstrap parameter to false to use the entire data set to build each tree which led to better classification results in initial experiments. } reveals the respective feature indices it requires to solve the classification task. Those can easily be mapped to electrodes on the EEG cap, and hence help to find the brain regions where activity varied across the different reading tasks.

 We extracted the $k$ most informative features to distinguish between NR and AR according to our Random Forest implementation. 
We conducted experiments over three different values for $k$ (i.e., $k \in [5, 15, 30]$), and scrutinized performances of different $k$-dimensional embeddings as input for an LSTM \cite{hochreiter1997lstm} to classify sentences in the respective tasks. This notably reduced the feature space from 4200 dimensions to $n \times m \times k \in {[15, 45, 90]}$, where $n = 1, m = 3, k = {[5, 15, 30]}$.

We computed $t$-test bootstrapping for each electrode in the four general frequency bands between NR and AR to inspect which electrodes show higher power spectra in which task (see Figure~\ref{fig:Figure_1}). All electrodes in the left temporal cortex, which is responsible for both language comprehension and production, that show significantly higher power spectra for AR compared to NR, were included in the $k$ most predictive features to solve the binary classification task through Random Forest. This means that Random Forest assigned more importance to brain signals that are enhanced for AR. We suspect that stronger EEG activity in these (fronto-)temporal areas of the human brain 
are due to higher cognitive load for AR compared to NR in language comprehension and production areas. However, further investigation must go into this line of research to draw definite conclusions. 

\subsection{Classifying task and session using reduced EEG features}

As as sanity check, we test the reduced EEG features in a simple, binary classification task to verify that we isolated the cognitive processing of text, and removed noise. We, therefore, use the reduced features to classify NR and AR, as well as whether signals were part of the first or the second half of Task 1. The latter was done to rule out the possibility that differences in brain signals between tasks occurred merely as a function of daytime. 

\paragraph{LSTM architecture} We use a vanilla LSTM\footnote{implemented end-to-end in PyTorch \cite{paszke2017automatic}.} with 1 hidden layer, 50 hidden units per layer, a layer dropout rate of 0.5, and an Adam optimizer \cite{kingma2014adam} with the default learning rate of $\alpha = 0.001$. Since the main goal was to predict the two different reading tasks, we minimized binary cross-entropy loss through mini-batch training with a batch size of 32. 

\begin{equation}
    \overrightarrow{\mathbf{h}}_{t} =\mathrm{LSTM}\left(\overrightarrow{\mathbf{h}}_{t-1}, \mathbf{x}_{t}\right), t=1, \cdots,|x|
\end{equation}

\noindent LSTM denotes the LSTM function \cite{hochreiter1997lstm}, h$_t$ is the hidden state at time step $t$, and $x_t$ represents the word input at current time step, where $x$ was embedded in $k$-dimensional EEG space ($x_t \in \mathbb{R}^{k}$).

\begin{table}[t]
\centering
\begin{footnotesize}
\begin{tabularx}{\linewidth} {@{}l|XXX|XXX@{}}
\toprule
&\multicolumn{3}{c}{\textsc{Validation}}&\multicolumn{3}{c}{\textsc{Test}}\\
\textsc{Task} & 15 &   45  & 90  &  15 &   45 & 90\\ 
\midrule
NR--AR & 100 & 100 & 93.8 & 100 & 98.4& 85.9 \\
Ses1--Ses2 & 48.4 & 57.8 & 50.5 & 45. & 50. & 48.8 \\
\bottomrule
\end{tabularx}
\end{footnotesize}
\caption{LSTM binary classification accuracies in \% with EEG word embeddings of different dimensions for two different tasks on both development and test set. Embedding dimensions were extracted through Random Forest tree splits for each task individually.}
\label{tab:Table_1}
\end{table}

\begin{figure}[h!]
\centering
\includegraphics[width=0.4\textwidth]{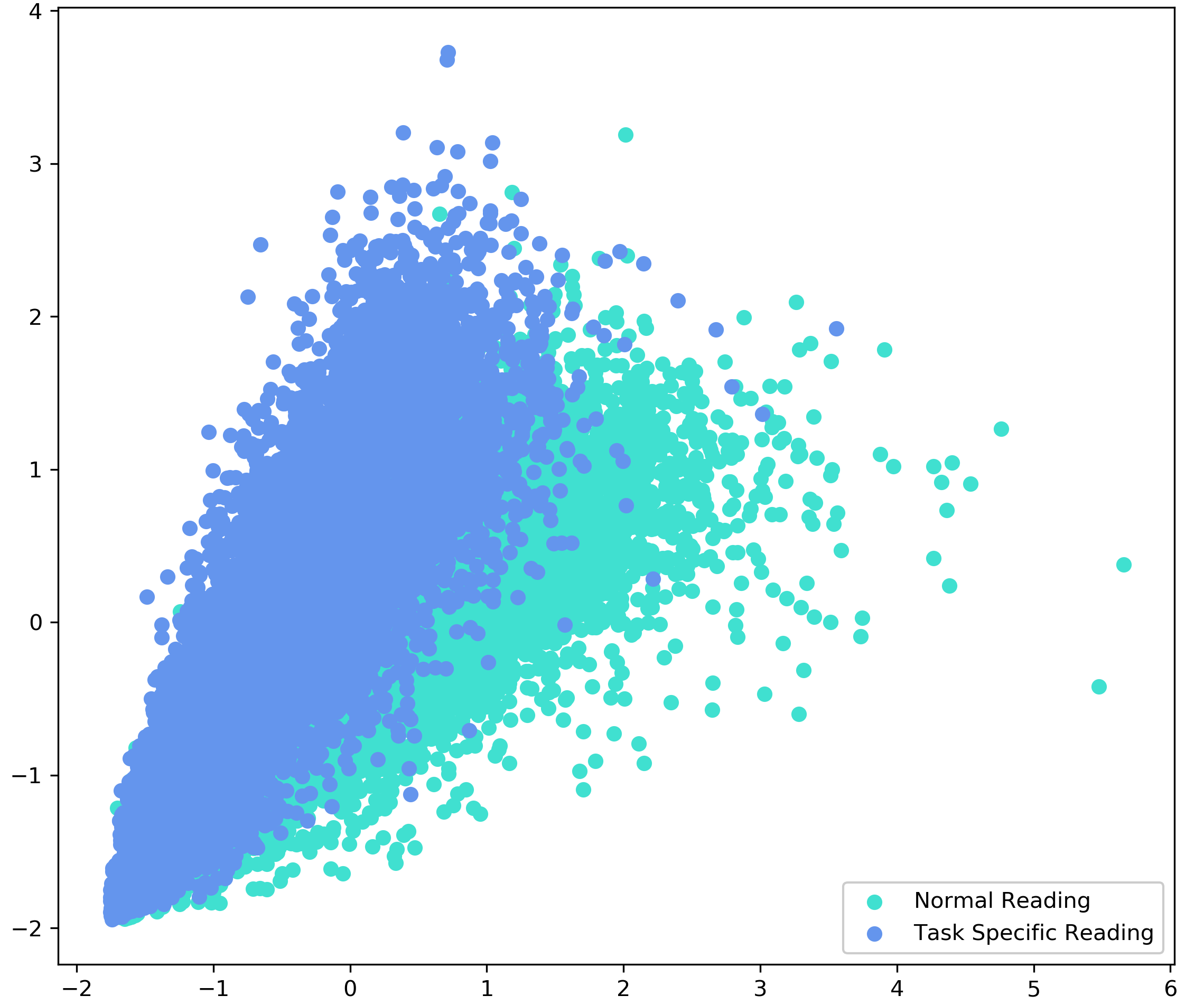}
\caption{Visualisation of the most informative feature for classifying task. Words embedded in two-dimensional brain space through Random Forest tree splits.}
\label{fig:Figure_2}
\end{figure}

The results in Table \ref{tab:Table_1} show that NR and AR appear to result in different brain signals and can thus be easily classified into two distinct classes in line with Figure~\ref{fig:Figure_1}. We experimented with different values for $k$ and found lower values of $k$ to be better at task classification. In contrast, session could not be classified into its respective days. It is, therefore, likely that our features capture a great amount of cognitive text processing signals and little noise.
We visualize each word that appeared in sentences in NR or AR respectively by the two most useful features (found by the Random Forest) in Figure~\ref{fig:Figure_2}. The plot shows that even in 2D space, EEG features well reflect the differences between NR and AR.

\section{Sequence labeling with EEG attention}

EEG activity can 
be used as a supervision, or guidance, signal for the attention function in neural networks. In the following section, we outline our attention-based model for sequence classification, present the task-specific datasets, and explain how we transform EEG to attention scores.

\subsection{EEG features to scalar}



Attention scores are scalar values that weigh their respective hidden word representations accordingly. To regularize attention, we were thus required to further reduce each $k$-dimensional word vector $x_t$ represented through EEG activity to a single dimension. We experimented with different approaches such as averaging across all $k$ electrodes or taking the maximum EEG power spectrum per $x_t$. Initial experiments on the development set revealed that taking the mean over a $k$ dimensional word vector leads to notably worse results than max-pooling. This might be due to information loss of brain activity as a result of averaging.

\begin{table*}[ht]
\centering

\begin{tabular} {@{}llrrrr@{}}
\toprule
\textsc{Task} &\textsc{Source}&\multicolumn{2}{c}{\textsc{Train}}&\multicolumn{1}{c}{\textsc{Val}}&\multicolumn{1}{c}{\textsc{Test}}\\
& & $n$ sents &\% positive   &  $n$ sents &  $n$ sents \\ 
\midrule
Relation Detection & SemEval 2010 &  8,096 & 19.3 & 1,361 &  1,372  \\
Relation Detection & Wikipedia & 1,733 &10.0 &  361   & 354  \\
NE Detection & Ontonotes 5.0 & 89,389 & 29.7 &  11,289 &  11,318 \\ 
\bottomrule
\end{tabular}

\caption{Overview of the data sets.}
\label{tab:Table_4}
\end{table*}

\paragraph{Domain specific scores} The final attention scores were computed through taking the maximum electrode value per $k$ dimensional word embedding for each frequency band. $k$ was one of [$15$, $45$, $90$] for concatenated embeddings, and one of [$5$, $15$, $30$] for embeddings per frequency domain. To yield values within the range [0, 1), we normalized each EEG attention score by the maximum attention score of the respective sentence. We observed that dividing each normalized attention score by some small constant $e$ leads to better performance. We assume this is due to the fact that EEG attention scores are somewhat peaky prior to dividing by $e$. Thus, the computation is as follows:

\begin{align}
    \begin{split}
        \mathbf{a}_{t}^{i} ={}& \mathrm{max}(\mathbf{{x}_{t} \in \mathbb{R}^{k}}), t=1, \cdots,|x|\\
    \end{split}\\
    \begin{split}
        \mathbf{a}_{t}^{i} ={}& \frac{\mathbf{a}^{i}_{t}}{\mathrm{max}(\mathbf{a}^{i})}, t=1, \cdots,|x| \\
    \end{split}\\
    \begin{split}
        \mathbf{a}_{t}^{i} ={}& \frac{\mathbf{a}^{i}_{t}}{e}, t=1, \cdots,|x|\\
    \end{split}
\end{align}

\noindent where $x_t$ denotes a word representation embedded in $k$-dimensional EEG space at time step $t$ for a sentence $i$. To compute $\mathbf{a}^i$ we did not exploit concatenated EEG embeddings but used isolated EEG embeddings for each of the three frequency domains $ \alpha, \beta, \theta $. Therefore, $k$ was one of ${[5, 15, 30]}$. The constant $e$ was set to $2$. The computed EEG attention scores served as inputs for our multi-task sequence classification model to supervise attention in the auxiliary task. Hence, final attention weights $\alpha^{i}$ were yielded through passing the EEG attention scores through the softmax function. The latter computation, however, happened automatically during training and was not done externally.
\begin{equation}
    \alpha{^i} = \operatorname{softmax}\left(\mathbf{a}^{i}\right)
\end{equation}

\begin{figure}[t]
\centering
\begin{subfigure}{0.51\textwidth}
    \centering
    \includegraphics[width=0.9\textwidth]{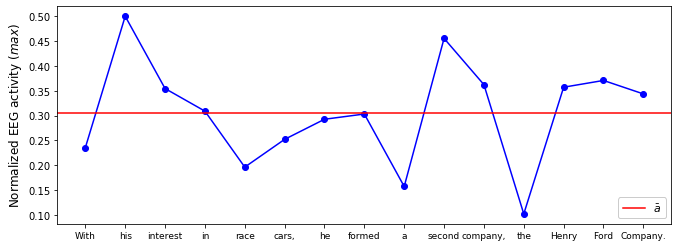}
    \caption{Normal Reading (NR)}
\end{subfigure}
\begin{subfigure}{0.51\textwidth}
    \centering
    \includegraphics[width=0.9\textwidth]{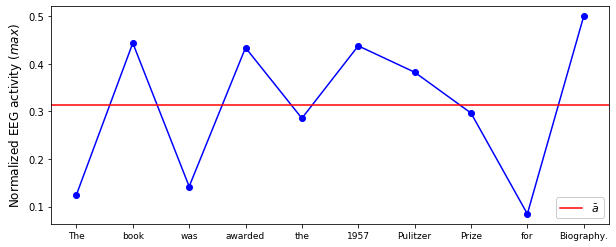}
    \caption{Annotation Reading (AR)}
\end{subfigure}
\caption{Max-pooled EEG attention scores $\mathbf{a}_{i}$ (\textit{averaged over all 12 participants}) for two example sentences read in Normal Reading (NR) and Annotation Reading (AR) respectively.}
\label{fig:Figure_3}
\end{figure}

\noindent Max-pooled EEG attention scores for two example sentences read in NR and AR respectively are depicted in Figure~\ref{fig:Figure_3}. We observe that brain activity transitions between words are smoother in NR compared to AR. We observe that for AR, in particular, there are higher activations for words relating to the relation \textsc{award}.

\subsection{Model}
The model is an adaptation of \newcite{rei2018zero} leveraged by \newcite{barrett2018sequence}.\footnote{\url{https://github.com/coastalcph/
Sequence_classification_with_
human_attention}} It is a biLSTM architecture that jointly learns the model parameters and the attention function by alternating training \cite{Luong:ea:16}, much related to multi-task learning  \cite{Dong:ea:15,Soegaard:Goldberg:16}. The inputs are token-level labelled sequences of EEG scalars to learn the attention function (auxiliary task) and a set of sentence-level labelled sequences for training the model parameters (main task). 

If the data point is from the main task, we perform normal training and model parameter update through comparing the model's class prediction ($\widehat{y}$) against the true label ($y$) on the sentence level. 

\begin{equation}
L_{sent}=\sum_{i}\left(y^{i}-\widehat{y}^{i}\right)^{2}
\end{equation}

If the data point is sampled from the EEG corpus, however, we do not update model parameters. We only modify the attention weights by minimizing the squared error between the EEG value and the attention score as described below.

\begin{equation}
L_{tok}=\sum_{i} \sum_{t}\left(\mathbf{a}^{i}_{t}-\widehat{\mathbf{a}}^{i}_{t}\right)^{2}
\end{equation}

The model has no access to EEG signals during testing. 

\subsection{Experiments}
Recall from Section \ref{sec:data} that participants read sentences from the Wikipedia relation extraction corpus \cite{culotta2006integrating}.
We employ three widely used Relation Extraction and NER benchmark data sets for English against baseline models without supervised attention and models whose attention was either supervised through eye-tracking data as in \newcite{barrett2018sequence} or word frequencies computed on the British National Corpus \cite{kilgarriff1995bnc}. 
We perform binary classification and adapt all datasets as described below to obtain sentence-level labels. As such, the main task was a $k$-vs.-the-rest binary sentence classification task.
Overall statistics about the data sets are displayed in Table~\ref{tab:Table_4}.

\subsubsection{Relation detection}

\begin{table*}[ht]
\begin{footnotesize}
\begin{tabularx}{\textwidth} {@{}l|XXX|XXX||XXr@{}}
\toprule
&\multicolumn{3}{c}{\textsc{SemEval 2010}}&\multicolumn{3}{c}{\textsc{Wiki}} &\multicolumn{3}{c}{\textsc{Ontonotes}}\\
\textsc{Attention} & Precision &   Recall &  $F$1 &  Precision &   Recall &       $F$1 &  Precision &   Recall &       $F$1\\ 
\midrule
baseline             &   80.03 &  63.21 &  70.56  &    54.44 &    55.00 &  54.67 & 88.90 &  64.46 &  74.72 \\
BNCFreqInv           &    78.30 &  58.00 &  66.52 &    61.39 &    \textbf{60.00} &  60.64  &    91.56 &  67.38 &  77.61\\
MeanFixCont          &    79.59 &  60.36 &  68.62 &   59.99 &    58.00 &  58.75 &    92.21 &  66.59 &  77.33 \\
\midrule
$5$k NR ($\alpha$)   &    77.90 &  \textbf{65.29} &  70.98  &    58.76 &    51.00 &  54.38 &    90.41 &  \textbf{67.84} &  77.51\\
$5$k NR ($\theta$)    &    79.70 &  61.21 &  69.14 &  \textbf{65.47} & 57.00 &  \textbf{60.74} &    91.02 &  67.18 &  77.28 \\
$15$k NR ($\alpha$)  &    79.66 &  63.93 &  70.91  &    54.58 &    51.00 &  52.52 &     91.25 &  67.36 &  77.50\\
$15$k NR ($\beta$)  &    78.58 &  64.43 &  70.79   &    61.09 &    57.00 &  58.82 &    91.85 &  67.00 &  77.47 \\
$15$k NR ($\theta$)  &     79.91 &  60.86 &  69.04 &    58.67 &    55.00 &  56.65 &   91.66 &  67.10 &  77.47\\
$30$k NR ($\alpha$)  &      77.88 &  64.57 &  70.53  &    52.83 &    49.00 &  50.77 &     91.04 &  67.68 &  \textbf{77.63}  \\
$30$k NR ($\theta$)  &     \textbf{80.32} &  61.14 &  69.33  &    56.88 &    53.00 &  54.64 &    \textbf{92.25} &  66.68 &  77.40 \\
\midrule
$5$k AR ($\alpha$)  &    79.66 &  59.64 &  68.18  &    60.25 &    49.00 &  53.80 &    91.19 &  66.84 &  77.13 \\
$5$k AR($\theta$)  &     79.55 &  63.00 &  70.29  &    56.42 &    49.00 &  52.42  &    91.01 &  67.01 &  77.18  \\
$15$k AR ($\alpha$) &     79.02 &  64.57 &  71.04 &    56.70 &    \textbf{60.00} &  57.75 &    90.97 &  67.17 &  77.23  \\
$15$k AR ($\beta$)   &    79.15 &  63.79 &  70.60 &    57.16 &    51.00 &  53.60 &    90.96 &  67.00 &  77.16  \\
$15$k AR ($\theta$) &    79.14 &  64.43 &  71.01  &    53.78 &    52.00 &  52.73 &   90.57 &  67.30 &  77.20  \\
$30$k AR($\alpha$) &    79.47 &  62.00 &  69.63  &    60.63 &    54.00 &  56.71 &   91.67 &  66.92 &  77.36  \\
$30$k AR ($\theta$) &    79.96 &  64.50 &  \textbf{71.34} &   59.63 &    52.00 &  55.16 &   91.03 &  66.87 &  77.10  \\

\bottomrule
\end{tabularx}
\end{footnotesize}
\caption{Relation detection and named entity detection. Results in \%. Best scores per metric are displayed in bold face. All scores are averaged over five random seeds.}
\label{tab:Table_2}
\end{table*}

\paragraph{SemEval 2010}

We used the SemEval 2010 Task 8 data set that defines the task as a multi-way classification of semantic relations between pairs of nominals \cite{hendrickx2009semeval}. The data set contains the following nine distinct relations: \textsc{Cause-Effect}, \textsc{Instrument-Agency}, \textsc{Product-Producer}, \textsc{Content-Container}, \textsc{Entity-Origin}, \textsc{Entity-Destination}, \textsc{Component-Whole}, \textsc{Member-Collection} and \textsc{Message-Topic}. Each sentence that contained the relations \textsc{Entity-Origin} or \textsc{Entity-Destination} served as a positive example of this relation. We have chosen those relations due to their higher frequency compared to other relations. Positive examples were tested against sentences that consisted of one of the remaining seven relations.

\paragraph{Wikipedia}
The data set provided by \citet{culotta2006integrating} contains Wikipedia articles labeled with 53 relation types. Since part of this dataset is included in ZuCo we filtered those sentences. 
We chose the sentences including the most frequent relation \textsc{job title} as positive examples.

\subsection{Named entity detection}

\paragraph{Ontonotes 5.0}
We use the four CoNLL-2003 NER labels PER, LOC, ORG, MISC in the Ontonotes 5.0 data set \cite{weischedel:ea:13} as positive examples.

\section{Results}
 
 \paragraph{SemEval 2010}
 The results are presented in Table~\ref{tab:Table_2} and we observe that EEG attention scores clearly help to solve the task. The best EEG-augmented model, $30$k AR ($\theta$), is better than all baselines. The most notable improvements are mainly due to a higher recall. Precision scores appear to be similar across all models although BNC word frequency augmented model show slightly lower precision than the rest.

\paragraph{Wikipedia}
The $F$1 score of the best EEG-augmented model, $5$k NR ($\theta$), outperformed all baselines (see Table~\ref{tab:Table_2}). The BNC-augmented model provides a strong baseline, which could be explained through both the significantly larger number of data points available for BNC word frequencies and due to the fact that word frequencies highly correspond to entities which are crucial to link entities.  

\paragraph{Ontonotes}
 The results show that all EEG augmented models outperform the baseline by a small margin (see Table~\ref{tab:Table_2}). The performance improvement for recall is again notable. Precision scores appear to be compareable across all models but the baseline. The best model is $30$k NR ($\alpha$).

\section{Discussion}
Human brain activity appears to help machine attention in attending toward the most crucial words in a sentence. However, there are differences between the exploited EEG attention scores - measured as the improvement in performance on a particular task. This is dependent on both frequency domain and reading task of the EEG signals, as well as the number of features captured in the embeddings. In general, we observe smaller performance gains for NER compared to Relation Extraction. We suppose this is due to NER results being fairly strong in general for English language data  \cite{DBLP:conf/naacl/LampleBSKD16,hollenstein2019entity}. Hence, additional support through cognitive data cannot enhance performance much. 

\paragraph{Reading tasks} Brain activity extracted from sentences read in NR appear to be more useful to detect named entities compared to EEG signals from AR. This is not surprising since in an NR setting participants read sentences without any additional task to perform (see \ref{sec:data}), and therefore read each sentence until its end \cite{hollenstein2018zuco}. On the other hand, brain signals distilled from AR guide the model to detect relations in a given sentence more effectively. This might be due to the fact that the readers were required to search for the respective relation while reading the sentence. Thus, participants drew particular attention to the decisive words that form the relation (Figure~\ref{fig:Figure_3}).

\paragraph{Frequency domains}
Performance enhancement differed depending on the frequency domain from which EEG attention scores were extracted. Both $\theta$  and $\alpha$ frequency bands show more useful signals and lead to better performance compared to $\beta$ across all tasks. Lower frequency bands such as $\theta$ (4-8 Hz) and $\alpha$ (8.5-13 Hz)  are linked to cognitive control \cite{williams2019thinking} and attentiveness \cite{klimesch2012alpha}, respectively. This might explain why brain signals from those domains are particularly useful to guide machine attention. Higher frequency domains such as $\beta$ (13.5-30 Hz) and $\gamma$ (30.5-49.5 Hz), however, are linked to motor activities \cite{pogosyan2009boosting} and enhanced emotional responses \cite{li2009emotion,oathes2008worry}, which explains why EEG power spectra in AR increase with the hertz rate (see Figure~\ref{fig:Figure_1}), but are less useful to supervise machine attention than brain signals from lower frequency bands (Table~\ref{tab:Table_2}).

\paragraph{Dimensions}
We tested EEG attention scores that were max-pooled over individual frequency domain embeddings of different dimensionality. Overall, the attention scores distilled from 15- and 30-dimensional embeddings carry slightly more informative signals than attention scores extracted from 5-dimensional embeddings (see Table~\ref{tab:Table_2}). The difference, however, is marginal, and for the Wikipedia relation detection, max-pooled attention scores over 5-dimensional $\theta$ embeddings outperform all other supervision signals. Contrary, to classify the word-level EEG signals of sentences into their respective reading tasks, lower-dimensional embeddings lead to better performance than higher-dimensional embeddings (see Table~\ref{tab:Table_1}). The latter is not surprising since low-dimensional embeddings contain those EEG signals that differ the most between NR and AR. The higher the dimensionality of the embeddings, the more EEG signals are captured that differ less between the two reading tasks.

\paragraph{Improvements despite limited data}
What is compelling is the fact that we exploited little EEG data to create the attention scores - merely 300 sentences for NR and 407 sentences for AR, averaged over 12 participants. Both ET and BNC frequency attention scores have $>10$ times more sentences to train the auxiliary task of the model. Even with such few EEG data samples, useful signals could be extracted that help neural networks to understand language in a similar manner as the human brain does. We assume that more data will lead to even higher performance gains. We plan studies to investigate the latter.

\section{Conclusions}
We presented the first study that leverages EEG activity to inform machine attention about language processing mechanisms of the human brain. This is compelling for two reasons: First, we successfully isolated the text processing signals from noisy EEG data, which considerably reduced its dimensions. Second, we demonstrated that even a small number of EEG data points from human readers can benefit multi-task neural models for sequence classification. Note that the extracted attention scores may be exploited in various neural architectures that employ any form of attention to solve NLP tasks. Third, we showed that downstream performance varies as a function of both cognitive load and EEG frequency domains. This might have decisive implications about which EEG signals are to inject into neural models. We suspect that more data will provide deeper, and more thorough insights into the latter avenue.

\section*{Acknowledgements}
We would like to thank Wolfgang Ganglberger, Martin Hebart, Joachim Bingel, and Desmond Elliot for fruitful comments on earlier versions of the paper.

\bibliography{references}
\bibliographystyle{acl_natbib}

\end{document}